\documentclass[runningheads]{llncs}
\usepackage[T1]{fontenc}
\usepackage{multirow}
\usepackage{booktabs}
\usepackage{graphicx}
\usepackage{amsmath}
\begin{document}
%

\title{From Detection to Mitigation: Addressing Gender Bias in Chinese Texts via Efficient Tuning and Voting-Based Rebalancing}


\author{Chengyan Wu\inst{1,2,*} \and
Yiqiang Cai\inst{1,2,*} \and
Yufei Cheng\inst{3}\and
Yun Xue\inst{1,2,\dagger}}
\footnotetext[1]{* Equal contribution.}
\footnotetext[2]{${\dagger}$ Corresponding author.}
\authorrunning{F. Author et al.}

\institute{Guangdong Provincial Key Laboratory of Quantum Engineering and Quantum Materials, School of Electronic Science and Engineering (School of Microelectronics), South China Normal University, Guangdong, China \and Guangdong Provincial Key Laboratory of Intelligent Information Processing, Guangdong, China \and School of Business, Yangzhou University, Yangzhou, China 
\\
\email{\{chengyan.wu,yiqiangcai,xueyun\}@m.scnu.edu.cn, chengyufei67@163.com}
}

\maketitle              
\begin{abstract}
This paper presents our team's solution to Shared Task 7 of NLPCC-2025, which focuses on sentence-level gender bias detection and mitigation in Chinese. The task aims to promote fairness and controllability in natural language generation by automatically detecting, classifying, and mitigating gender bias. To address this challenge, we adopt a fine-tuning approach based on large language models (LLMs), efficiently adapt to the bias detection task via Low-Rank Adaptation (LoRA). In terms of data processing, we construct a more balanced training set to alleviate class imbalance and introduce heterogeneous samples from multiple sources to enhance model generalization. For the detection and classification sub-tasks, we employ a majority voting strategy that integrates outputs from multiple expert models to boost performance. Additionally, to improve bias generation detection and mitigation, we design a multi-temperature sampling mechanism to capture potential variations in bias expression styles. Experimental results demonstrate the effectiveness of our approach in bias detection, classification, and mitigation. Our method ultimately achieves an average score of 47.90\%, ranking fourth in the shared task.

\end{abstract}
\section{Introduction}
With the widespread application of artificial intelligence technologies in real-world scenarios such as recruitment, question answering, translation, and recommendation, the issue of potential gender bias in models has increasingly attracted attention. Some researchers argue that identifying and preventing harmful gender bias is of significant societal importance \cite{blodgett2020language}. Gender bias refers to a preference for or prejudice against one gender over another \cite{moss2012science}. It manifests in various components of natural language processing (NLP) systems, including training data, pre-trained models, and algorithms \cite{zhao2018gender,bolukbasi2016man,garg2018word}. Bias present in any of these components can lead to gender-biased predictions in NLP systems, sometimes even amplifying existing biases \cite{zhao2017men}. The propagation of gender bias through NLP algorithms poses the risk of reinforcing harmful stereotypes, thereby affecting downstream applications. This phenomenon can have serious real-world consequences; for example, there is concern that automated resume screening systems may exhibit a preference for male applicants when gender is the only distinguishing factor.

Since most gender bias in real life is expressed and disseminated through text, developing models capable of detecting and mitigating gender bias in textual content is particularly critical \cite{zhang2023corgi}. Researchers have proposed training models using gender-swapped corpora \cite{lu2020gender}, though such methods heavily rely on the quality of the underlying models. To address this issue, some studies have adopted supervised training on manually annotated biased corpora, but these approaches tend to focus primarily on lexical-level bias \cite{webster2018mind}. Moreover, most existing methods have been developed for English and lack adaptation to Chinese \cite{costa2019analysis}. With the recent rapid development of large language models (LLMs), such as GPT, Claude, and Deepseek, it has become feasible to employ LLMs for gender bias-related tasks. Trained on massive corpora, LLMs possess a deep understanding of complex linguistic phenomena and are capable of identifying subtle and implicit gender-discriminatory expressions.

In this work, we propose an LLM-based framework for detecting and mitigating gender bias in Chinese text. Our approach leverages the CORGI-PM dataset \cite{zhang2023corgi} and employs Low-Rank Adaptation (LoRA) to efficiently fine-tune the model for bias detection tasks. For the classification subtask, we address label imbalance through data recombination and improve performance via ensemble prediction using majority voting from multiple expert models. Furthermore, to enhance the diversity of generated de-biased texts, we design a multi-temperature sampling strategy that captures stylistic variations in biased expressions.

The contributions of this work are as follows:
\begin{itemize}
    \item We propose a unified framework for Chinese gender bias probe and mitigation, utilizing parameter-efficient fine-tuning (LoRA) to adapt large language models across all subtasks with minimal computational overhead.
    \item To address label imbalance and ambiguity in the classification subtask, we design a data recombination approach combined with majority voting from multiple expert models, significantly enhancing classification robustness.
    \item For the mitigation subtask, we introduce a multi-temperature sampling strategy to increase the diversity and stylistic variation of bias-corrected outputs, improving both linguistic richness and bias coverage.
\end{itemize}



\section{Related Work}
\subsection{Text Style Transfer}
Previous approaches to Text Style Transfer (TST) can be categorized based on the type of training data used, namely parallel and non-parallel data methods. Parallel data-based methods typically adopt a sequence-to-sequence framework \cite{sutskever2014sequence}, which can be implemented using architectures such as Transformer \cite{vaswani2017attention} and LSTM \cite{memory1997sepp}. However, due to the scarcity of parallel data in real-world scenarios, these methods are often impractical. Consequently, the majority of TST research focuses on non-parallel data. Non-parallel data methods primarily fall into three categories: prototype editing \cite{li2018delete}, disentanglement-based approaches \cite{zhang2018style}, and pseudo-parallel corpus construction \cite{jin2019imat}.

Recently, LLMs have transformed the landscape of TST in natural language processing. In exploring how to leverage LLMs for TST effectively, Reif et al. \cite{reif2021recipe} enhanced LLM performance through few-shot prompting, while Liu et al. \cite{liu2024adaptive} proposed a dynamic prompt generation method to guide the model in producing text with the desired style. Although prompt engineering has proven useful, the performance of LLMs is highly sensitive to prompt variations \cite{zhu2023promptrobust}, which may lead to instability in output quality. In contrast, using fine-tuned LLMs for TST has been shown to yield more consistent results. Motivated by these insights, this work adopts a hybrid approach, combining prompt-based guidance with further fine-tuning of the model to perform bias-mitigated style transfer for gender discrimination. This method integrates the strengths of both strategies, enabling the model to produce high-quality outputs while maintaining greater stability.
\subsection{Text Classification}
Text classification refers to the task of assigning predefined categories to text sequences. In recent years, with the rapid development of social networks, blogs, and forums, as well as the continuous expansion of online academic resource repositories, research on text classification has become increasingly important. Early approaches to text classification mostly relied on manually constructed features \cite{mikolov2013efficient,pennington2014glove}, and used machine learning algorithms for classification. For example, Chen et al. \cite{chen2017comparative} employed simple logistic regression methods for classifying textual information, while Hinrich et al. \cite{schutze2008introduction}, considering computational efficiency in machine learning, adopted a Naive Bayes classifier for document classification.

With the rise of deep learning models, text classification has entered a new stage of development. CNNs and RNNs are representative architectures in deep learning. Chen et al. \cite{chen2015convolutional} applied CNNs to sentence-level text classification tasks, extracting local features using multiple convolutional kernels. Given that CNNs are less effective in capturing long-range dependencies in text, Zhou et al. \cite{zhou2015c} proposed a hybrid CNN-RNN model for long-text classification tasks.With the advancement of LLMs, researchers have also begun using these models for text classification tasks. Devlin et al. \cite{devlin2019bert} introduced the BERT model, which unified many NLP downstream tasks and has been widely applied to text classification. Raffel et al. \cite{raffel2020exploring} proposed the T5 framework, which reformulates text classification as a text generation task. Inspired by these developments, this study adopts an instruction-tuned large language model to determine whether a given sentence contains gender-biased content.

\subsection{LoRA Framework}

LLMs have achieved remarkable success in natural language processing tasks \cite{chu2024qwen2}, demonstrating powerful emergent abilities \cite{wei2022chain}. However, directly training LLMs requires substantial computational resources. To address this, various fine-tuning methods have been proposed to reduce resource demands, with an increasing number of researchers adopting lightweight fine-tuning techniques to adapt LLMs to different downstream tasks.

For the task of gender bias detection and mitigation, this work employs LoRA for fine-tuning. LoRA works by keeping the original model weights frozen and introducing trainable low-rank decomposition matrices as adapter modules at each layer of the model. Compared to the base LLM, these adapter modules contain significantly fewer trainable parameters, enabling efficient adaptation to new tasks with limited data. Compared to earlier tuning methods such as Prefix-Tuning \cite{li2021prefix} and P-Tuning v2 \cite{liu2021p}, LoRA offers advantages including fewer modified layers and easier training, making it a more efficient and scalable fine-tuning approach.

\section{Methods}
In this section, we first define the task, followed by a detailed description of our proposed method. The architecture of our model is illustrated in Fig \ref{fig:overall_model}. Overall, we propose a fine-tuning strategy for LLMs that integrates efficient adaptation with class-balancing techniques. We also describe the majority voting strategy employed in our approach.
\subsection{Problem Statement}
The task focuses on detecting, classifying, and mitigating gender bias in Chinese texts, and is structured into three subtasks:

\begin{figure*}[htbp]
\centering
\includegraphics[width=0.85\textwidth]{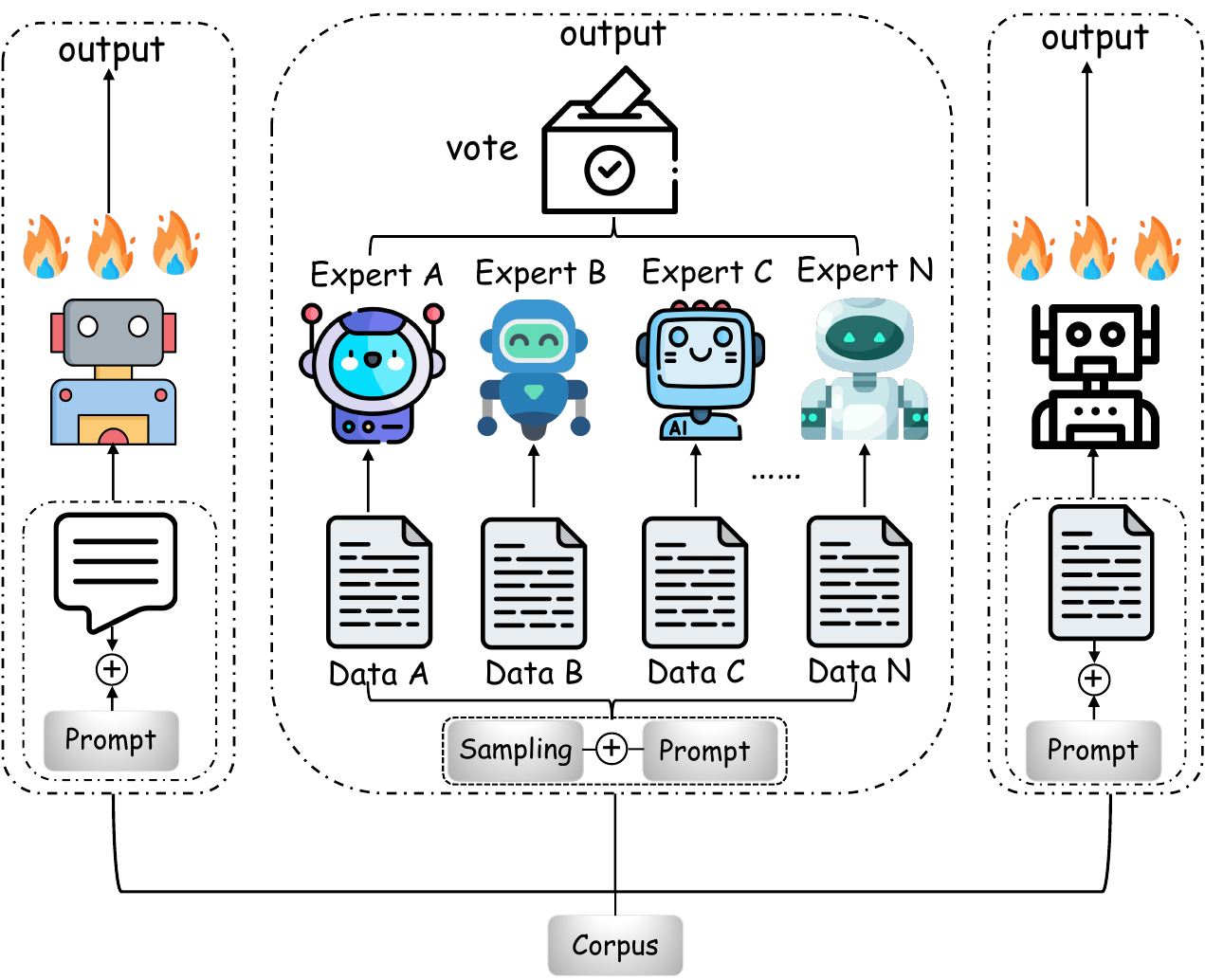}
\caption{The overall framework of the proposed method. From left to right, it illustrates the specific implementations of subtasks 1, 2, and 3.}
\label{fig:overall_model}
\end{figure*}

\emph{\textbf{Bias Detection.}} Given a sentence, the model is required to determine whether it contains gender bias. Each input sentence is labeled as either biased (B) or non-biased (N). The output is a boolean value: \textbf{true} if the sentence contains gender bias, and \textbf{false} otherwise. 

\emph{\textbf{Bias Classification.}} For sentences identified as biased, the model must further classify the bias into one or more predefined categories: (1) AC: Gender-stereotyped activities and career choices. (2) DI: Gender-stereotyped descriptions and inductions. (3) ANB: Expressed gender-stereotyped attitudes, norms, and beliefs.
The output is a multi-hot vector indicating the presence of each bias type (e.g.,\texttt{[1, 0, 0]} indicates the presence of the AC bias type.).

\emph{\textbf{Bias Mitigation.}} Given a biased sentence, the goal is to generate a revised version that mitigates or removes the gender bias while preserving the original semantics. This may involve rewriting, substituting biased expressions, or adding neutralizing commentary.

\subsection{Prompt Setting}
\begin{figure*}[htbp]
    \centering
    \includegraphics[width=0.65\textwidth]{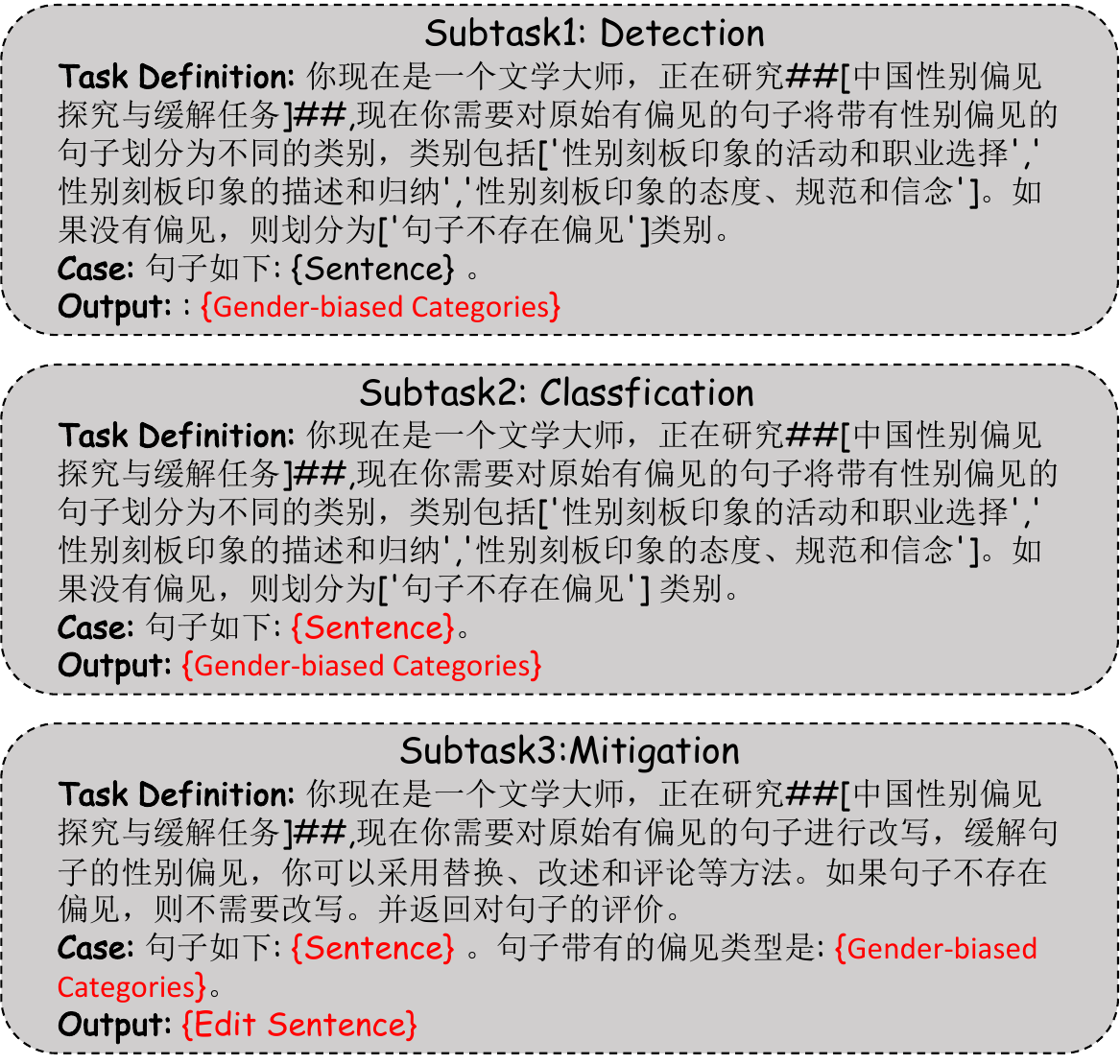}
    \caption{The prompts used in the LLMs.}
    \label{fig:prompt}
\end{figure*}

Fig \ref{fig:prompt} illustrates the prompt design used in our framework. For all three subtasks, we adopt a unified prompt structure that consists of role-playing, task description, and test examples. Specifically, for subtasks 2 and 3, we convert the original multi-hot vectors into textual descriptions of bias types, enabling the model to interpret and process them in natural language form. In particular, the bias categories in subtask 3 are obtained based on the predictions from subtask 2 and subsequently incorporated into the model input.


\subsection{Class balance and Lora Fine-Tuning}
For subtask 1, we perform efficient LoRA fine-tuning \cite{wang2023lora} on the model using 4,172 biased data instances. For subtask 2, the training set consisting of 21,418 unbiased instances is randomly split into five subsets. Each subset is then randomly combined with the biased data at approximately a 1:1 ratio to construct diverse training sets. In addition to preserving the completeness of the data, we also create a combined dataset that includes all biased and unbiased instances without any splitting. For subtask 3, we generate texts with different styles by varying the temperature settings.



\subsection{Majority Voting}
In text classification tasks within the field of natural language processing, voting mechanisms are commonly employed as ensemble strategies to aggregate predictions from multiple models and improve overall performance. For instance, Onan and Wu et al. \cite{onan2016multiobjective,wu2023medical,wu2023model} proposed a weighted voting scheme for sentiment analysis, where model outputs are combined with different weights to enhance classification accuracy. Similarly, Yin et al. \cite{yin2024crisissense} leveraged prompt engineering techniques with large language models (LLMs) and adopted a majority voting strategy for multi-label classification of social media texts, achieving notable improvements in model performance.

The core advantage of such methods lies in their ability to allow each individual expert to focus on different subspaces or characteristics of the training data. This fosters diversity among models while enhancing the robustness of the overall system. Inspired by ensemble learning \cite{agbesi2024mutcelm,cang2024albert}, our work incorporates a six-expert voting system, where each expert is fine-tuned on a differently restructured version of the training dataset. This diversification results in experts having distinct predictive preferences and biases. However, due to class imbalance or the presence of rare categories, some experts may struggle with specific classes, making single-expert predictions unreliable.

To address this issue, we design a majority voting mechanism: when more than three out of six experts agree on a prediction, we take this majority decision as the final output. In cases where no majority is formed, the system retains the original confidence scores for downstream processing. This strategy effectively mitigates overfitting to specific training features by individual experts, and enhances the generalization capability and stability of the model, particularly in challenging bias-type classification tasks.

\section{Experiments}

\subsection{Datasets}
The CORGI-PM dataset \cite{zhang2023corgipmchinesecorpusgender} consists of Chinese sentences collected from diverse sources such as social media and news commentaries, with a focus on gender bias in the Chinese socio-cultural context. The dataset contains approximately 329,000 sentences, each annotated with high-quality labels by annotators with backgrounds in linguistics and gender studies, using an annotation scheme specifically designed for gender bias. The annotations include the type of bias, the target of the bias, and fine-grained subcategory labels, enabling deeper understanding and analysis of gender bias in text. The entire dataset is divided into training, validation, and test sets to ensure robust and fair evaluation. Table \ref{tab:data_statistic} presents more detailed statistics of the dataset.

\begin{table*}[t]
\centering
\caption{Size of sentences for the official datasets.}
\begin{tabular}{c|c|c|c}
\hline
\textbf{} & \textbf{\# biased}  & \textbf{\# non-biased} & \textbf{Total}  \\
\hline
 Train& 4172 & 21418 & 25590 \\
  Dev& 516 & 516 & 1032 \\
\hline
   & \textbf{subtask1} & \textbf{subtask2\&3} & \textbf{Total} \\
   \hline
    Test& 200 & 200 & 400 \\
\hline
\end{tabular}

\label{tab:data_statistic}
\end{table*}

\subsection{Training Details}

 We use the \texttt{Qwen2.5-7B-Instruct}\footnote{https://huggingface.co/Qwen/Qwen2.5-7B-Instruct} model, with a learning rate of $3\times10^{-4}$, batch size of 8, and 4 training epochs. The inference temperature is set to 0.1 for subtasks 1 and 2, while for subtask 3, the temperature values range over $\{0.01,\ 0.1,\ 0.3\}$ (As shown in Table \ref{tab:Task123 three temperature result}). Our experiments are conducted using the \texttt{PyTorch} deep learning framework, and we adopt a mainstream LoRA fine-tuning strategy. Notably, we employ the \texttt{llama-factory}\footnote{https://github.com/hiyouga/LLaMA-Factory} tool to facilitate model training.

\begin{table*}[t]
\centering
\caption{Top-4 official overall result of NLPCC2025 Task7.}
\begin{tabular}{c|c|c|c|c|c}
\hline
\textbf{Rank} & \textbf{System Name}  & \textbf{Task1 Score} & \textbf{Task2 Score} & \textbf{Task3 Score}& \textbf{Average Score} \\
\hline
 1& ZZU-nlp & 0.850 & 0.646 & 0.294& 0.597\\
  2& YNU-HPCC & 0.714 & 0.509 & 0.293& 0.505\\
   3& Prompt & 0.712 & 0.505 & 0.288& 0.502\\
    4& Cloud Lab & 0.720 & 0.453 & 0.265& 0.479\\
\hline
\end{tabular}
\label{tab:Top-4 official result}
\end{table*}

\begin{table*}[t]
\centering
\caption{Precision, Recall and F1 score of experiments for subtask 1.}
\begin{tabular}{c|c|c|c|c}
\hline
\textbf{Rank} & \textbf{System Name}  & \textbf{Precision} & \textbf{Recall} & \textbf{F1} \\
\hline
 1& ZZU-nlp & 0.798 & 0.910 & 0.850 \\
2 & Cloud Lab & 0.566 & 0.990 & 0.720 \\
3 & YNU-HPCC & 0.645 & 0.800 & 0.714 \\
4 & Prompt & 0.552 & 1.000 & 0.712 \\
\hline
\end{tabular}

\label{tab:Task1 result}
\end{table*}

\begin{table*}[t]
\centering
\scriptsize
\setlength\tabcolsep{3pt}
\caption{Performance of each subtask at different temperatures.}
\begin{tabular}{c|c|c|c}
\hline
\textbf{Temperature} & \textbf{Task1 Score}  & \textbf{Task2 Score} & \textbf{Task3 Score}  \\
\hline
0.01& 0.720 & 0.453 & 0.394  \\
0.1 & 0.720 & 0.453 & 0.393  \\
0.3 & 0.720 & 0.453 & 0.392  \\
\hline
\end{tabular}
\label{tab:Task123 three temperature result}
\end{table*}

\subsection{Evaluation Metrics}
Tasks related to gender bias are typically evaluated using different metrics depending on the subtask: detection and classification tasks are assessed using the F1 score, while mitigation tasks are evaluated based on the average of BLEU, METEOR, and ROUGE-L F1 scores.

BLEU \cite{papineni2002bleu} is an automatic evaluation metric based on n-gram precision matching. In this metric, BP refers to the brevity penalty, which is used to penalize excessively short outputs, and $p_n$ represents the n-gram precision score.

\begin{equation}
BLEU = BP \cdot \exp\left( \sum_{n=1}^{N} w_n \cdot \log p_n \right)
\end{equation}


METEOR \cite{banerjee2005meteor} is an evaluation metric designed to assess the quality of generated text, primarily developed to address the limitations of early metrics such as BLEU. The core formula is shown below, where F denotes the weighted harmonic mean, and Penalty represents the penalty term.

\begin{equation}
     METEOR = F_{\text{mean}} \cdot (1 - \text{Penalty}) \\
\end{equation}

ROUGE-L \cite{lin2004rouge} is a variant of the ROUGE evaluation metric. The corresponding formula is shown below, where F1 denotes the harmonic mean of precision and recall, and LCS(X, Y) represents the length of the longest common subsequence between sequences X and Y.

\begin{equation}
R_{\text{LCS}} = \frac{\text{LCS}(X, Y)}{m}
\end{equation}
\begin{equation}
P_{\text{LCS}} = \frac{\text{LCS}(X, Y)}{n}
\end{equation}
\begin{equation}
F_{\text{LCS}} = \frac{(1 + \beta^2) \cdot R_{\text{LCS}} \cdot P_{\text{LCS}}}{R_{\text{LCS}} + \beta^2 \cdot P_{\text{LCS}}}
\end{equation}





\subsection{Results and Analysis}
Tables \ref{tab:Top-4 official result}, \ref{tab:Task1 result}, \ref{tab:Task2 result}, and \ref{tab:Task3 result} present the average overall scores and subtask-specific scores of the top four teams in terms of overall performance.

In subtask 1, our method ranks second in the final leaderboard, demonstrating the effectiveness of our fine-tuning strategy. We perform efficient fine-tuning based on the Qwen model, focusing on how different hyperparameter configurations affect performance in bias detection. In particular, our optimization in LoRA design and learning rate scheduling confirms the practical value of our efficient parameter search strategy.

In subtask 2, there exists a performance gap between our approach and the top-performing team. We adopt a data compression and reorganization strategy, restructuring the original dataset according to different bias categories and training multiple sub-models to perform classification. The final prediction is produced via a multi-expert voting mechanism. While this method improves the model’s ability to identify common bias types, we reuse the same hyperparameters from subtask 1 without task-specific tuning, which limits the model’s upper-bound performance. Furthermore, our approach to class imbalance considers only the ratio between biased and non-biased data, while overlooking the imbalance within the biased categories. We plan to address this by incorporating loss re-weighting, oversampling, and data augmentation strategies to enhance the model’s recognition of long-tail categories.

In subtask 3, we also fine-tune the model and focus on the impact of generation strategies on the quality of bias mitigation. We fix most parameters and vary the temperature during generation to produce more diverse and expressive rewrites. Due to resource and submission constraints, we explore only three temperature values, which may not cover the optimal search space. Additionally, we retain the same training configuration used in subtask 1 without further tuning for text generation quality. In future work, we plan to introduce evaluation metrics such as BLEU, BERTScore, and MAUVE, and implement automatic temperature selection or multi-pass generation with reranking to improve the diversity and controllability of the outputs. Notably, the competition limits us to three final submissions, and we do not submit our best-performing configuration, indicating that our model still has room for further improvement.



\begin{table*}[t]
\centering
\caption{Precision, Recall and F1 score of experiments for subtask 2.}
\begin{tabular}{c|c|c|c|c}
\hline
\textbf{Rank} & \textbf{System Name}  & \textbf{Precision} & \textbf{Recall} & \textbf{F1} \\
\hline
1 & ZZU-nlp & 0.543 & 0.576 & 0.646 \\
2 & YNU-HPCC & 0.463 & 0.565 & 0.509 \\
3 & Prompt & 0.496 & 0.513 & 0.505 \\
4 & Cloud Lab & 0.510 & 0.409 & 0.453 \\
\hline
\end{tabular}

\label{tab:Task2 result}
\end{table*}



\begin{table}[h]
 \caption{BLEU, METEOR and ROUGE-L score of experiments for subtask 3.}
    \centering
    \begin{tabular}{c|c|c|c|c|c|c|c}
        \hline
       \multirow{2}{*}{\textbf{Rank}} &\multirow{2}{*}{\textbf{System Name}} & \multirow{2}{*}{\textbf{BLEU}} & \multirow{2}{*}{\textbf{METEOR}} & \multicolumn{3}{c|}{\textbf{ROUGE-L}} & \multirow{2}{*}{\textbf{Average}} \\ 
       \cline{5-7}
        & & & & \textbf{Precision} & \textbf{Recall} & \textbf{F1} & \\ \hline
        1 & ZZU-nlp & 0.013 & 0.417 & 0.464 & 0.452 & 0.452 & 0.294 \\ 
        2 & YNU-HPCC & 0.013 & 0.414 & 0.472 & 0.448 & 0.453 & 0.293 \\ 
        3 & Prompt & 0.011 & 0.418 & 0.430 & 0.462 & 0.434 & 0.288 \\ 
        4 & Cloud Lab & 0.009 & 0.391 & 0.367 & 0.461 & 0.394 & 0.265 \\ \hline
    \end{tabular}
    \label{tab:Task3 result}
\end{table}

\section{Conclusion}
In this paper, we propose a unified framework for gender bias detection and mitigation, with a primary focus on the challenge of class imbalance. To address this, we design a simple yet effective prompt-based fine-tuning approach. Specifically, we recombine biased and unbiased samples to construct multiple resampled subsets, each of which is used to train an expert model based on a pretrained backbone. The final prediction is obtained via majority voting across these experts, which enhances the model’s ability to recognize minority-class biases. In addition, we further improve performance through efficient hyperparameter tuning, including LoRA fine-tuning and temperature adjustments. Our team ranks fourth in the NLPCC-2025 Task 7 competition, demonstrating the effectiveness of the proposed approach in real-world scenarios.

\bibliographystyle{splncs04}
\bibliography{custom}

\begin{thebibliography}{10}
\providecommand{\url}[1]{\texttt{#1}}
\providecommand{\urlprefix}{URL }
\providecommand{\doi}[1]{https://doi.org/#1}

\bibitem{agbesi2024mutcelm}
Agbesi, V.K., Chen, W., Yussif, S.B., Ukwuoma, C.C., Gu, Y.H., Al-Antari, M.A.: Mutcelm: An optimal multi-textcnn-based ensemble learning for text classification. Heliyon  \textbf{10}(19) (2024)

\bibitem{banerjee2005meteor}
Banerjee, S., Lavie, A.: Meteor: An automatic metric for mt evaluation with improved correlation with human judgments. In: Proceedings of the acl workshop on intrinsic and extrinsic evaluation measures for machine translation and/or summarization. pp. 65--72 (2005)

\bibitem{blodgett2020language}
Blodgett, S.L., Barocas, S., Daum{\'e}~III, H., Wallach, H.: Language (technology) is power: A critical survey of" bias" in nlp. arXiv preprint arXiv:2005.14050  (2020)

\bibitem{bolukbasi2016man}
Bolukbasi, T., Chang, K.W., Zou, J.Y., Saligrama, V., Kalai, A.T.: Man is to computer programmer as woman is to homemaker? debiasing word embeddings. Advances in neural information processing systems  \textbf{29} (2016)

\bibitem{cang2024albert}
Cang, Y., Yang, W., Sun, D., Ye, Z., Zheng, Z.: Albert-driven ensemble learning for medical text classification. Journal of Computer Technology and Software  \textbf{3}(6) (2024)

\bibitem{chen2017comparative}
Chen, W., Xie, X., Wang, J., Pradhan, B., Hong, H., Bui, D.T., Duan, Z., Ma, J.: A comparative study of logistic model tree, random forest, and classification and regression tree models for spatial prediction of landslide susceptibility. Catena  \textbf{151},  147--160 (2017)

\bibitem{chen2015convolutional}
Chen, Y.: Convolutional neural network for sentence classification. Master's thesis, University of Waterloo (2015)

\bibitem{chu2024qwen2}
Chu, Y., Xu, J., Yang, Q., Wei, H., Wei, X., Guo, Z., Leng, Y., Lv, Y., He, J., Lin, J., et~al.: Qwen2-audio technical report. arXiv preprint arXiv:2407.10759  (2024)

\bibitem{costa2019analysis}
Costa-Juss{\`a}, M.R.: An analysis of gender bias studies in natural language processing. Nature Machine Intelligence  \textbf{1}(11),  495--496 (2019)

\bibitem{devlin2019bert}
Devlin, J., Chang, M.W., Lee, K., Toutanova, K.: Bert: Pre-training of deep bidirectional transformers for language understanding. In: Proceedings of the 2019 conference of the North American chapter of the association for computational linguistics: human language technologies, volume 1 (long and short papers). pp. 4171--4186 (2019)

\bibitem{garg2018word}
Garg, N., Schiebinger, L., Jurafsky, D., Zou, J.: Word embeddings quantify 100 years of gender and ethnic stereotypes. Proceedings of the National Academy of Sciences  \textbf{115}(16),  E3635--E3644 (2018)

\bibitem{jin2019imat}
Jin, Z., Jin, D., Mueller, J., Matthews, N., Santus, E.: Imat: Unsupervised text attribute transfer via iterative matching and translation. arXiv preprint arXiv:1901.11333  (2019)

\bibitem{li2018delete}
Li, J., Jia, R., He, H., Liang, P.: Delete, retrieve, generate: a simple approach to sentiment and style transfer. arXiv preprint arXiv:1804.06437  (2018)

\bibitem{li2021prefix}
Li, X.L., Liang, P.: Prefix-tuning: Optimizing continuous prompts for generation. arXiv preprint arXiv:2101.00190  (2021)

\bibitem{lin2004rouge}
Lin, C.Y.: Rouge: A package for automatic evaluation of summaries. In: Text summarization branches out. pp. 74--81 (2004)

\bibitem{liu2024adaptive}
Liu, Q., Qin, J., Ye, W., Mou, H., He, Y., Wang, K.: Adaptive prompt routing for arbitrary text style transfer with pre-trained language models. In: Proceedings of the AAAI Conference on Artificial Intelligence. vol.~38, pp. 18689--18697 (2024)

\bibitem{liu2021p}
Liu, X., Ji, K., Fu, Y., Tam, W.L., Du, Z., Yang, Z., Tang, J.: P-tuning v2: Prompt tuning can be comparable to fine-tuning universally across scales and tasks. arXiv preprint arXiv:2110.07602  (2021)

\bibitem{lu2020gender}
Lu, K., Mardziel, P., Wu, F., Amancharla, P., Datta, A.: Gender bias in neural natural language processing. Logic, language, and security: essays dedicated to Andre Scedrov on the occasion of his 65th birthday pp. 189--202 (2020)

\bibitem{memory1997sepp}
Memory, L.S.T.: Sepp hochreiter and j{\"u}rgen schmidhuber. Neural Computation  \textbf{9}(8), ~1735 (1997)

\bibitem{mikolov2013efficient}
Mikolov, T., Chen, K., Corrado, G., Dean, J.: Efficient estimation of word representations in vector space. arXiv preprint arXiv:1301.3781  (2013)

\bibitem{moss2012science}
Moss-Racusin, C.A., Dovidio, J.F., Brescoll, V.L., Graham, M.J., Handelsman, J.: Science faculty’s subtle gender biases favor male students. Proceedings of the national academy of sciences  \textbf{109}(41),  16474--16479 (2012)

\bibitem{onan2016multiobjective}
Onan, A., Koruko{\u{g}}lu, S., Bulut, H.: A multiobjective weighted voting ensemble classifier based on differential evolution algorithm for text sentiment classification. Expert Systems with Applications  \textbf{62},  1--16 (2016)

\bibitem{papineni2002bleu}
Papineni, K., Roukos, S., Ward, T., Zhu, W.J.: Bleu: a method for automatic evaluation of machine translation. In: Proceedings of the 40th annual meeting of the Association for Computational Linguistics. pp. 311--318 (2002)

\bibitem{pennington2014glove}
Pennington, J., Socher, R., Manning, C.D.: Glove: Global vectors for word representation. In: Proceedings of the 2014 conference on empirical methods in natural language processing (EMNLP). pp. 1532--1543 (2014)

\bibitem{raffel2020exploring}
Raffel, C., Shazeer, N., Roberts, A., Lee, K., Narang, S., Matena, M., Zhou, Y., Li, W., Liu, P.J.: Exploring the limits of transfer learning with a unified text-to-text transformer. Journal of machine learning research  \textbf{21}(140),  1--67 (2020)

\bibitem{reif2021recipe}
Reif, E., Ippolito, D., Yuan, A., Coenen, A., Callison-Burch, C., Wei, J.: A recipe for arbitrary text style transfer with large language models. arXiv preprint arXiv:2109.03910  (2021)

\bibitem{schutze2008introduction}
Sch{\"u}tze, H., Manning, C.D., Raghavan, P.: Introduction to information retrieval, vol.~39. Cambridge University Press Cambridge (2008)

\bibitem{sutskever2014sequence}
Sutskever, I., Vinyals, O., Le, Q.V.: Sequence to sequence learning with neural networks. Advances in neural information processing systems  \textbf{27} (2014)

\bibitem{vaswani2017attention}
Vaswani, A., Shazeer, N., Parmar, N., Uszkoreit, J., Jones, L., Gomez, A.N., Kaiser, {\L}., Polosukhin, I.: Attention is all you need. Advances in neural information processing systems  \textbf{30} (2017)

\bibitem{wang2023lora}
Wang, X., Aitchison, L., Rudolph, M.: Lora ensembles for large language model fine-tuning. arXiv preprint arXiv:2310.00035  (2023)

\bibitem{webster2018mind}
Webster, K., Recasens, M., Axelrod, V., Baldridge, J.: Mind the gap: A balanced corpus of gendered ambiguous pronouns. Transactions of the Association for Computational Linguistics  \textbf{6},  605--617 (2018)

\bibitem{wei2022chain}
Wei, J., Wang, X., Schuurmans, D., Bosma, M., Xia, F., Chi, E., Le, Q.V., Zhou, D., et~al.: Chain-of-thought prompting elicits reasoning in large language models. Advances in neural information processing systems  \textbf{35},  24824--24837 (2022)

\bibitem{wu2023model}
Wu, C., Fang, W., Dai, F., Yin, H.: A model ensemble approach with llm for chinese text classification. In: China Health Information Processing Conference. pp. 214--230. Springer (2023)

\bibitem{wu2023medical}
Wu, C., Lin, Z., Fang, W., Huang, Y.: A medical diagnostic assistant based on llm. In: China Health Information Processing Conference. pp. 135--147. Springer (2023)

\bibitem{yin2024crisissense}
Yin, K., Liu, C., Mostafavi, A., Hu, X.: Crisissense-llm: Instruction fine-tuned large language model for multi-label social media text classification in disaster informatics. arXiv preprint arXiv:2406.15477  (2024)

\bibitem{zhang2023corgi}
Zhang, G., Li, Y., Wu, Y., Zhang, L., Lin, C., Geng, J., Wang, S., Fu, J.: Corgi-pm: A chinese corpus for gender bias probing and mitigation. arXiv preprint arXiv:2301.00395  (2023)

\bibitem{zhang2023corgipmchinesecorpusgender}
Zhang, G., Li, Y., Wu, Y., Zhang, L., Lin, C., Geng, J., Wang, S., Fu, J.: Corgi-pm: A chinese corpus for gender bias probing and mitigation (2023), \url{https://arxiv.org/abs/2301.00395}

\bibitem{zhang2018style}
Zhang, Z., Ren, S., Liu, S., Wang, J., Chen, P., Li, M., Zhou, M., Chen, E.: Style transfer as unsupervised machine translation. arXiv preprint arXiv:1808.07894  (2018)

\bibitem{zhao2017men}
Zhao, J., Wang, T., Yatskar, M., Ordonez, V., Chang, K.W.: Men also like shopping: Reducing gender bias amplification using corpus-level constraints. arXiv preprint arXiv:1707.09457  (2017)

\bibitem{zhao2018gender}
Zhao, J., Wang, T., Yatskar, M., Ordonez, V., Chang, K.W.: Gender bias in coreference resolution: Evaluation and debiasing methods. arXiv preprint arXiv:1804.06876  (2018)

\bibitem{zhou2015c}
Zhou, C., Sun, C., Liu, Z., Lau, F.: A c-lstm neural network for text classification. arXiv preprint arXiv:1511.08630  (2015)

\bibitem{zhu2023promptrobust}
Zhu, K., Wang, J., Zhou, J., Wang, Z., Chen, H., Wang, Y., Yang, L., Ye, W., Zhang, Y., Gong, N., et~al.: Promptrobust: Towards evaluating the robustness of large language models on adversarial prompts. In: Proceedings of the 1st ACM Workshop on Large AI Systems and Models with Privacy and Safety Analysis. pp. 57--68 (2023)

\end{thebibliography}




\end{document}